\documentclass[10pt,twocolumn,letterpaper]{article}

\usepackage{iccv}
\usepackage{times}
\usepackage{epsfig}
\usepackage{graphicx}
\usepackage{amsmath}
\usepackage{amssymb}
\usepackage{multirow}
\usepackage{pifont}
\usepackage{dblfloatfix}

\usepackage[accsupp]{axessibility}  

\usepackage{xcolor}

\usepackage{url}

\usepackage[breaklinks=true,bookmarks=false]{hyperref}

\iccvfinalcopy 

\def\iccvPaperID{****} 

\ificcvfinal\fi
\begin{document}

\title{MV-DeepSDF: Implicit Modeling with Multi-Sweep Point Clouds\\
for 3D Vehicle Reconstruction in Autonomous Driving}



\author{Yibo Liu\textsuperscript{1,2}, Kelly Zhu\textsuperscript{1,3}, Guile Wu\textsuperscript{1}, Yuan 
Ren\textsuperscript{1},  Bingbing Liu\textsuperscript{1}, Yang Liu\textsuperscript{1}, Jinjun Shan\textsuperscript{2}\\
\textsuperscript{1}Huawei Noah’s Ark Lab,
\textsuperscript{2}York University,
\textsuperscript{3}University of Toronto\\
{\tt\small \{yorklyb,jjshan\}@yorku.ca, kellyk.zhu@mail.utoronto.ca }\\
{\tt\small\{guile.wu, yuan.ren3, liu.bingbing, yang.liu9\}@huawei.com}
}


\maketitle
\ificcvfinal\thispagestyle{empty}\fi

\begin{abstract}
Reconstructing 3D vehicles from noisy and sparse partial point clouds is of great significance to autonomous driving.
Most existing 3D reconstruction methods cannot be directly applied to this problem because they are elaborately designed to deal with dense inputs with trivial noise.
In this work, we propose a novel framework, dubbed MV-DeepSDF,
which estimates the optimal Signed Distance Function (SDF) shape representation from multi-sweep point clouds
to reconstruct vehicles in the wild.
Although there have been some SDF-based implicit modeling methods,
they only focus on single-view-based reconstruction, resulting in low fidelity.
In contrast, we first analyze multi-sweep consistency and complementarity in the latent feature space
and propose to transform the implicit space shape estimation problem into an element-to-set feature extraction problem.
Then, we devise a new architecture to extract individual element-level representations and
aggregate them to generate a set-level predicted latent code.
This set-level latent code is an expression of the optimal 3D shape in the implicit space,
and can be subsequently decoded to a continuous SDF of the vehicle.
In this way, our approach learns consistent and complementary information among multi-sweeps for 3D vehicle reconstruction.
We conduct thorough experiments on two real-world autonomous driving datasets (Waymo and KITTI) to demonstrate the superiority of our approach over state-of-the-art alternative methods both qualitatively and quantitatively.
\end{abstract}

\begin{figure}[t!]
	\centering
	\includegraphics[width=3.3in]{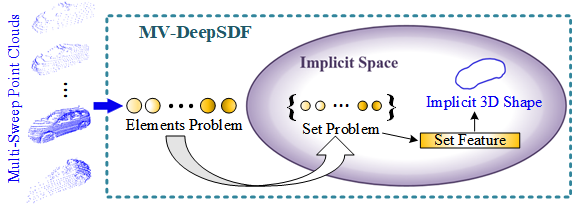}
	\caption{An illustration of the motivation of the proposed approach.
	Our approach takes multi-sweep point clouds as input and
	simplifies 3D vehicle reconstruction from multi-sweeps into an element-to-set feature extraction problem.
	In this way, we infer an optimal estimation of the 3D shape described in the abstract implicit space
	for 3D vehicle reconstruction in autonomous driving.}
	\label{illu}
\end{figure}

\begin{figure*}
	\centering
	\includegraphics[width=17.5cm]{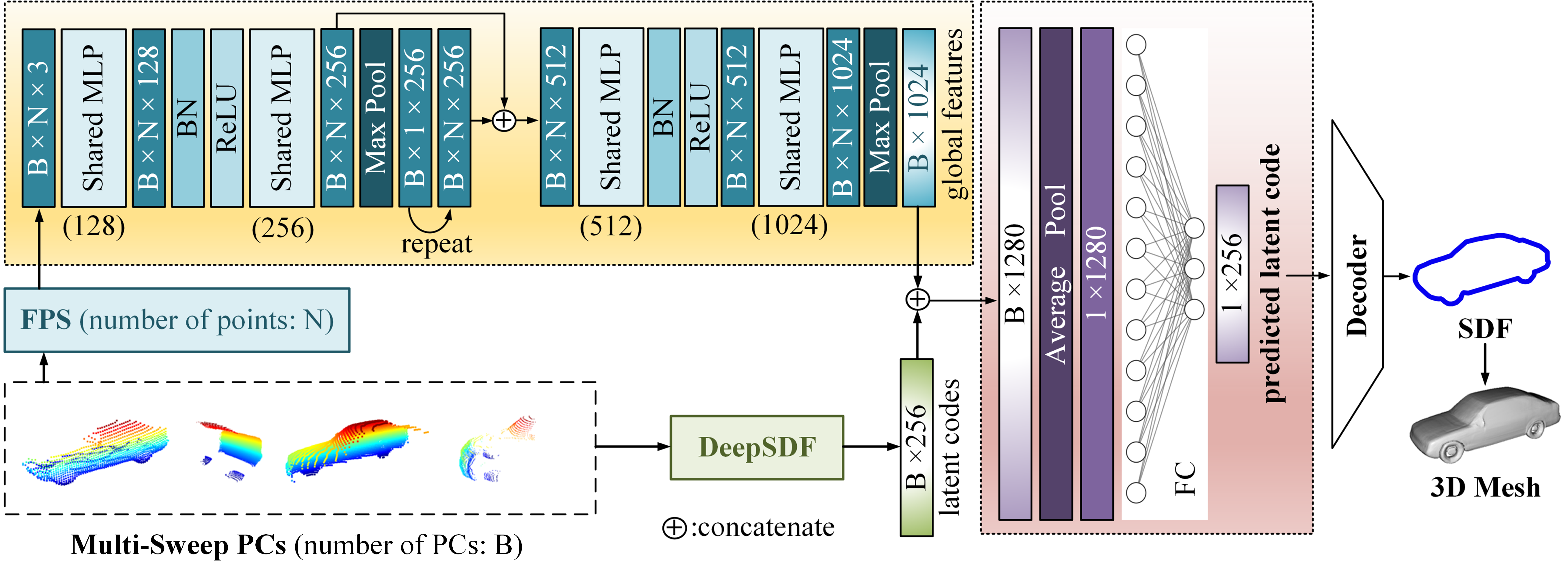}
	\caption{The framework of the proposed MV-DeepSDF.
	Farthest Point Sampling (FPS) \cite{pointnet2} is applied to the raw point clouds for pre-processing
	and DeepSDF \cite{deepsdf} is employed to generate a latent code for each observation.
	We then extract global features from the standardized point clouds (refer to yellow block) and concatenate the
	global features with the latent codes as \emph{element-level representations}.
	Next, the element-level representations are transformed into a
	\emph{set-level predicted latent code} (refer to red block).
	Finally, we employ a pre-trained DeepSDF decoder \cite{deepsdf}  to project the predicted latent code to the SDF
	of the vehicle in 3D space and recover the 3D mesh.
	Note that the decoder used in the final step (refer to white trapezoid) is identical to that of DeepSDF (refer to green block).}
	\label{network}
\end{figure*}

\section{Introduction}
3D vehicle reconstruction from sparse and partial point clouds is a fundamental need in the autonomous driving
industry \cite{mending,weakly}.
It aims to infer the 3D structure of vehicles at arbitrary resolutions in the wild,
which is of great significance to many downstream tasks in autonomous driving.
Despite the many 3D reconstruction methods \cite{nerf,dvgo,neat,trans,neus,depth,render4,depth2},
most of them focus on image-based inputs \cite{nerf,dvgo,neat,trans,neus}, dense point cloud inputs \cite{render4,deepsdf,csdf}, or their combination \cite{depth,depth2,depth3} and thus, cannot be directly applied to this problem. \par

Recently, \cite{mending} has shown promising performance in applying implicit modeling to tackle this problem.
Contrary to explicit modeling methods \cite{pointnet, pointnet2,pointr,pcn,grnet,voxel1,voxel2,mesh1,mesh2}
that directly represent 3D object shape structure with points, voxels, or meshes,
implicit modeling maps the 3D shape to a low-dimensional latent space and
learns the projection from the latent space to a continuous function that describes the 3D shape.
This presents two significant advantages:
1) the 3D shape can be stored as a low-dimensional memory-saving latent 
code \cite{deepsdf,csdf,mending,deepls};
2) the trained network outputs a continuous function in the 3D space which supports mesh extraction at any resolution \cite{deepsdf}.
To this end, we also choose to employ implicit modeling in our approach.

However, previous point cloud-based implicit modeling methods \cite{deepsdf,csdf,mending}
mainly focus on recovering 3D shapes from a single-view partial point cloud
and fail to leverage the multi-sweep information of vehicles.
As a result, when noise or annotation errors exist in an individual sweep,
single-view methods usually produce low fidelity results.
In real-world datasets (\eg, \cite{waymo,kitti}), multi-sweep point clouds are usually available
and contain richer shape information because of the various viewing angles offered by multiple observations \cite{weakly}.
Although there are some existing 3D reconstruction methods from multi-view point clouds \cite{weakly,render4,licp,gicp},
implicit modeling with multi-sweep point clouds remains an unsolved problem.

In this work, we propose a novel framework, dubbed MV-DeepSDF,
to exploit multi-sweep point clouds and implicitly generate high-fidelity 3D reconstructions of vehicles for autonomous driving.
An illustration of the motivation for our proposed approach is depicted in Figure \ref{illu}.
Specifically, to shed light on the importance of exploring multi-sweep point clouds for 3D vehicle reconstruction,
we first analyze multi-sweep consistency and complementarity in the latent feature space.
We then propose to consider the problem of shape estimation in the implicit space as an element-to-set feature extraction problem,
where a set represents a collection of multi-sweeps and an element represents a sparse and partial point cloud.
Next, we devise a new architecture to simultaneously extract a global feature and latent code for each element in the multi-sweep
and concatenate them as element-level representations.
These element-level representations are then transformed into a set-level predicted latent code with the aid of
average pooling and mapping.
This set-level latent code is an optimal estimation of the 3D shape described in the abstract implicit space,
which we subsequently decode to a continuous SDF of the vehicle with a pre-trained DeepSDF decoder \cite{deepsdf}
and recover the 3D mesh from the SDF. The \textbf{contributions} of this work are threefold:
\begin{itemize}
\item We analyze multi-sweep consistency and complementarity in the latent feature space and
	transform the problem of shape estimation in the implicit space into an element-to-set feature extraction problem.
	This simplifies 3D vehicle reconstruction from multi-sweep point clouds into an optimal estimation of the 3D shape described in the abstract implicit space.
\item We propose a novel MV-DeepSDF framework (see Figure \ref{network}) for implicit modeling with multi-sweep point clouds.
	A new architecture is constructed to extract the element-level representations
	and generate the set-level predicted latent code.
\item We qualitatively and quantitatively demonstrate the superior performance of our approach over the state-of-the-art alternative methods through extensive experiments on real-world datasets \cite{waymo,kitti}.
\end{itemize}

\section{Related Work}
\label{rel}
\subsection{Explicit Modeling-Based 3D Reconstruction}
Conventional 3D explicit modeling can be classified into three categories,
namely point-based \cite{pointnet, pointnet2,pointr,pcn,grnet}, voxel-based \cite{voxel1,voxel2}, and mesh-based \cite{mesh1,mesh2}.
Point-based methods, such as Point Completion Network (PCN) \cite{pcn}, GRNet \cite{grnet},
and PoinTr \cite{pointr}, output a point cloud with limited resolution since the number of points is fixed
and not suitable for generating watertight surfaces due to lack of topology description.
Voxel-based methods \cite{voxel1,voxel2} describe volumes by subdividing the space into a 3D grid,
but they are memory and compute intensive, leading to slow training and low resolution representations \cite{deepsdf}.
Mesh-based methods only generate meshes with simple topologies \cite{mesh3}
given by a fixed reference template from the same object class and cannot guarantee watertight surfaces \cite{mesh5,mesh4}.
In contrast, we propose implicit modeling with multi-sweep point clouds in a novel MV-DeepSDF framework,
which estimates the optimal SDF shape representation for reconstructing 3D structures of vehicles in autonomous driving. 

\subsection{Implicit Modeling-Based 3D Reconstruction}
\label{related}
Implicit modeling \cite{occ,deepsdf,csdf,mending} uses function-based decision boundaries to implicitly define surfaces for 3D representations.
%
%
DeepSDF \cite{deepsdf} is a classical approach to implicit modeling.
It estimates the Signed Distance Function (SDF) of an object from a partial point cloud,
where the SDF specifies whether a querying position is inside or outside the surface of the object and the distance of the querying position from the surface.
There have been some improved versions \cite{csdf,mending,deepls,gifs} of DeepSDF,
%
but these methods only focus on 3D reconstruction from a single-view point cloud
and thus, generally produce low fidelity results when noise or annotation errors exist in an individual sweep.
In contrast, we propose a novel framework for implicit modeling with multi-sweep point clouds.
We analyze multi-sweep consistency and complementarity in the latent feature space and
propose to resolve this problem by an optimal estimation of the 3D shape described in the abstract implicit space.
Moreover, other implicit modeling-based methods, such as Neural Radiance Fields (NeRF) \cite{nerf},
Point-NeRF \cite{pointnerf}, Direct Voxel Grid Optimization \cite{dvgo}, NeuS \cite{neus}, NEAT \cite{neat}, and \cite{depth,depth2,depth3} exist,
but they all require image input and cannot deal with only LiDAR data.

\subsection{Multi-View 3D Reconstruction}
3D reconstruction from multi-view point clouds can be categorized into conventional approaches \cite{licp,gicp,tsdf}
and deep learning-based approaches \cite{weakly,render4,similar}.
Conventional geometric approaches, such as Iterative Closest Point (ICP), TSDF \cite{tsdf}, and \cite{licp,gicp},
only leverage multi-view observations geometrically, so the resulting completion is merely an aligned stack of partial point clouds
and cannot produce watertight shapes unless the sensor fully loops around the object.
In deep learning-based approaches, \cite{weakly} proposes a weakly-supervised framework 
to directly learn 3D shape completion from multi-view LiDAR sweeps, but its reconstruction result is not watertight.
\cite{render4} presents a shape completion framework using a multi-view depth map-based shape representation approach,
while \cite{similar} designs a network to use multiple partial point clouds encoded into the latent space to estimate the optimal shape of the object.
However, the output of these networks \cite{render4,similar} are point-based representations and in contrast with continuous field functions such as SDFs \cite{deepsdf}, lack continuity in 3D space.
Moreover,  some methods, such as \cite{samp,lm,online}, exist for reconstructing vehicle shapes from multiple observations,
but \cite{samp} relies on stereo images and motion priors to regularize the shape estimation of vehicles,
\cite{lm} requires multi-view camera images to build semantic maps which contain 3D vehicle shapes,
and \cite{online} requires inputs to follow a time sequence to track and reconstruct 3D objects.
In contrast, our approach simplifies 3D vehicle reconstruction from multi-sweep point clouds into an optimal estimation of the 3D shape described in the abstract implicit space,
which supports mesh extraction at any resolution and can deal with noise and annotation errors that exist in an individual sweep.

\section{Methodology}
\subsection{Preliminaries} \label{pre}
In this section, we introduce preliminaries regarding DeepSDF \cite{deepsdf} since our approach 
utilizes its framework for extracting the latent code and decoding the continuous SDF for 3D mesh generation.
Formally, when adopting the DeepSDF decoder into the shape completion task, points of partial point clouds are taken as given surface points
and the SDF value is defined as the distance from the querying point to the surface of the object.
Interior and exterior querying points are sampled along the normals of the surface points
and the SDF value is computed for each querying point.
Suppose the set of newly sampled interior and exterior points is denoted as $\mathcal{P_{M}}$, where $M$ is the number of points.
Define the  $m$th point in $\mathcal{P_{M}}$ as $\boldsymbol{x}_m$ and its corresponding SDF value as $s_m$.
In the domain of DeepSDF \cite{deepsdf}, a 3D shape can be represented by a latent code $\boldsymbol{z} \in \mathbb{R}^{256}$ in the latent space.
The complete shape is obtained by optimizing $\boldsymbol{z}$ via Maximum a Posterior estimation:
\begin{equation}
\hat{\boldsymbol{z}}=\frac{1}{\sigma^2}\|\boldsymbol{z}\|_2^2+\underset{\boldsymbol{z}}{\arg \min } \sum_{m=1}^M\mathcal{L}_{1}\left(f_\theta\left(\boldsymbol{z}, \boldsymbol{x}_m\right), s_m\right) \label{recon},
\end{equation}
where $\mathcal{L}_{1}$ represents the clamped $\mathcal{L}_{1}$ distance and $\sigma$ is a regularization hyperparameter.
Once the optimal latent code $\hat{\boldsymbol{z}}$ is determined, the complete 3D shape can be recovered by finding the zero-SDF-value isosurface through
the DeepSDF decoder $f(\cdot)$, which is defined as:
\begin{equation}
\hat{s}=f_\theta\left(\hat{\boldsymbol{z}}, \boldsymbol{x}\right) \label{sdf},
\end{equation}
where $\theta$ represents the parameters of the decoder, $\boldsymbol{x}\in \mathbb{R}^{3}$ denotes the 3D coordinates of the querying point, and $\hat{s}$ denotes the estimated SDF value given by $f(\cdot)$.
Here, the sign of $\hat{s}$, either positive or negative, indicates whether the point lies on the exterior or interior of the surface, respectively.
Thus, the surface of the object can be implicitly represented by the isosurface composed of points with zero SDF values.

Despite its simplicity and efficiency, DeepSDF only takes the given surface points into consideration when generating 3D shapes (\ie, Eq.~\eqref{recon}).
This can lead to suboptimal reconstruction results in areas not captured by the partial point clouds since DeepSDF will reconstruct arbitrarily based on prior knowledge learned during training.
For instance, referencing the ground truth of Case 1 in Figure \ref{different}, the reconstruction results of DeepSDF for Cases 2, 3, and 4 only demonstrate high fidelity in areas with well-captured surface points. The missing areas are reconstructed arbitrarily.
%
To resolve this problem, we analyze multi-sweep consistency and complementarity in the latent feature space
and propose implicit modeling with multi-sweep point clouds in MV-DeepSDF.

\subsection{Approach Overview}
As shown in Figure \ref{network}, there are three main steps in MV-DeepSDF, namely preprocessing, optimal latent code prediction, and 3D mesh extraction. \par
\textbf{1) Preprocessing.}
First, we carry out Farthest Point Sampling (FPS) \cite{pointnet2} to sample a fixed number of points on each raw partial point cloud
since our global feature extractor requires standardized point clouds as input.
For FPS, the number of centroids is set as 256.

\textbf{2) Optimal Latent Code Prediction.}
The post-FPS point clouds are used as input into the global feature extractor (yellow block in Figure \ref{network})
to extract global features for each individual point cloud.
Meanwhile, the original point clouds are used to extract latent codes for each partial point cloud through
the pre-trained DeepSDF (green block in Figure \ref{network}).
Then, as shown in the red block of Figure \ref{network},
the global features and latent codes are concatenated as element-level representations,
followed by an average pooling operation to aggregate the information into a single instance tensor.
Finally, this instance tensor is transformed into the predicted latent code by a fully-connected layer as the set-level representation.

\textbf{3) 3D Mesh Extraction.}
The predicted latent code is converted to an SDF using a pre-trained DeepSDF decoder \cite{deepsdf}.
The isosurface composed of querying points with zero SDF values represents the surface of the instance and the implicit surface can be rasterized to a 3D mesh using Marching Cubes \cite{marching}.
Since the SDF is a continuous function, it supports 3D mesh extraction at any resolution.

\subsection{Consistency and Complementarity Analysis} \label{analysis}

\paragraph{Background.}
In literature~\cite{multi,comple},
multi-view consistency refers to the consistency of describing the same instance from different viewpoint observations and multi-view complementarity refers to the complementary information of the same instance provided by different observations.
Existing methods mainly explore multi-view consistency and complementarity for 3D reconstruction with partial point clouds in explicit modeling \cite{gicp,licp}.

However, existing methods are not applicable for exploring multi-sweep/-view consistency and complementarity in implicit modeling because the abstract feature space operates differently than the explicit space.
In implicit modeling, point clouds are expressed as low-dimensional feature vectors (a.k.a. latent vectors) in the latent feature
space and the latent vectors are generated through a non-linear learning process that projects the 3D space to the feature space
through a deep neural network \cite{deepsdf,pointnet,pointnet2,pcn}. \par

\begin{figure}[t!]
	  \centering
	  \includegraphics[width=3.3in]{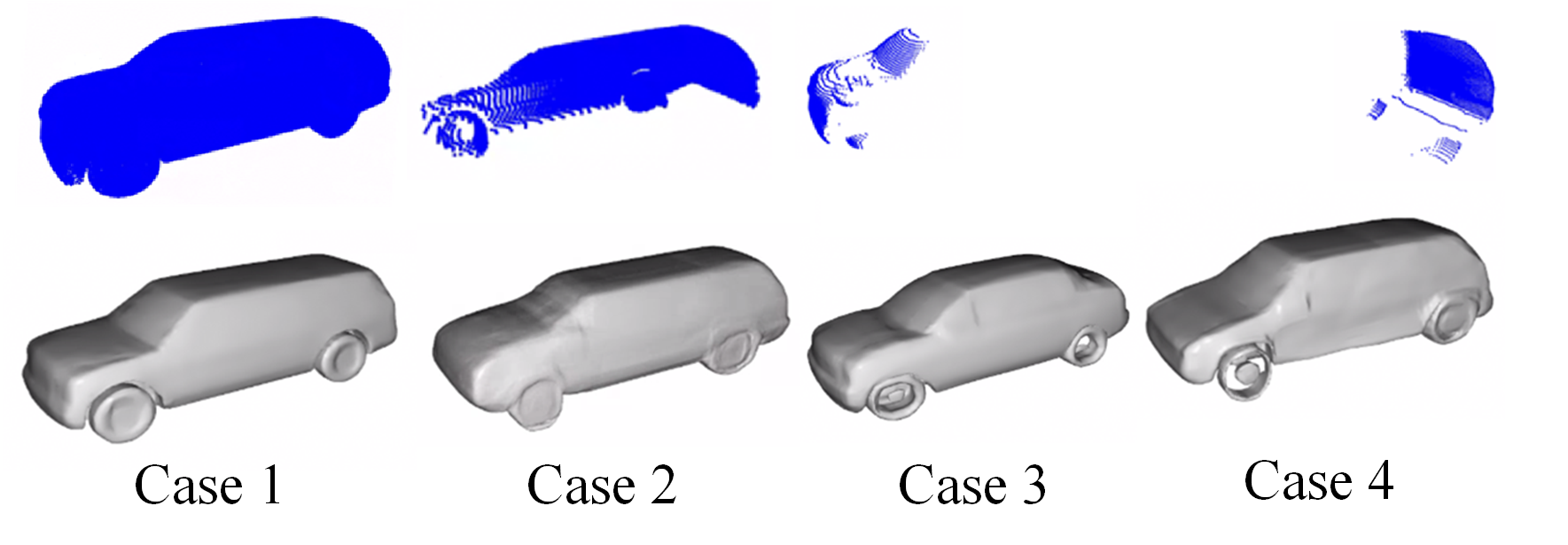}
	  \caption{The 3D reconstruction results of DeepSDF \cite{deepsdf} (row 2) using different  point clouds (row 1). All point clouds are sampled from the same CAD model of ShapeNetV2 \cite{shapenet}. Case 1 shows the complete dense point cloud used to train the DeepSDF decoder, while Cases 2, 3, and 4 show the partial point clouds.}
	  \label{different}
  \end{figure}

\paragraph{Theoretical Analysis.}
Despite these differences, we can still explore multi-sweep consistency and complementarity in the latent feature space
for implicit modeling.
Inspired by \cite{consist}, consider a total of $B$ observations. The latent code for the $i$th observation, denoted by $\boldsymbol{z}_{i}$, can be represented as:
\begin{equation}
	\boldsymbol{z}_{i} = \boldsymbol{z}_{i,c} + \boldsymbol{z}_{i,s} + \boldsymbol{e}_{i} \label{c1},
	\end{equation}
where $i=\{1,\cdots,B\}$,  $\boldsymbol{z}_{i,c}$ is the consistent component, $\boldsymbol{z}_{i,s}$ is the specific component,
and $\boldsymbol{e}_{i}$ is the error component.
In this work, $\boldsymbol{z}_{i,c}$ and $\boldsymbol{z}_{i,s}$ can be defined as follows:
\begin{equation}
 \boldsymbol{z}_{i,c}=  \boldsymbol{z}_{i}\cap \boldsymbol{z}_{gt} \quad \mathrm{and}
 \quad (\boldsymbol{z}_{i,s} + \boldsymbol{e}_{i}) = \boldsymbol{z}_{gt}- \boldsymbol{z}_{i}   \label{c2},
\end{equation}
where $\boldsymbol{z}_{gt}$ is the latent code corresponding to the ground truth shape and 
`$\cap$' and `$-$' represent the similarity and difference in information captured by the two feature vectors, respectively.
Now, consider any two consistent components $\boldsymbol{z}_{i,c}$ and $\boldsymbol{z}_{j,c}$.
In the first case where $\boldsymbol{z}_{i,c}\cap\boldsymbol{z}_{j,c}=\varnothing$, the two components contain completely different consistent information, which is complementary.
In the second case where $\boldsymbol{z}_{i,c}\cap\boldsymbol{z}_{j,c}\neq \varnothing$, but $ \boldsymbol{z}_{i,c} \not\subseteq \boldsymbol{z}_{j,c}$ and $ \boldsymbol{z}_{j,c} \not\subseteq \boldsymbol{z}_{i,c}$,
the two components contain some different information, which can be aggregated as complementary information to approach the ground truth.
In the final case where $\boldsymbol{z}_{i,c}\cap\boldsymbol{z}_{j,c}\neq \varnothing$, but $ \boldsymbol{z}_{i,c}\subseteq\boldsymbol{z}_{j,c}$ or $ \boldsymbol{z}_{j,c}\subseteq\boldsymbol{z}_{i,c}$, either $\boldsymbol{z}_{i,c}$ or $\boldsymbol{z}_{j,c}$ contains redundant information from the other component. While redundant, this will not adversely impact complementary aggregation.
%
%
Thus, to make the optimal $\hat{\boldsymbol{z}}$ approach the ground truth $\boldsymbol{z}_{gt}$,
it is desired to aggregate all consistent and complementary information among all individual features:
\begin{equation}
\hat{\boldsymbol{z}} = \boldsymbol{z}_{1,c}\cup\boldsymbol{z}_{2,c} \cdots\cup\boldsymbol{z}_{B,c}+\boldsymbol{p}  \label{c3},
\end{equation}
where `$\cup$' refers to the complementary aggregation of information among features and
$\boldsymbol{p}$ denotes the predicted information for regions not captured in the multi-sweeps
(\eg, the side of the vehicle not captured by the multi-sweeps as shown in the Cases 3 and 4 of Figure \ref{different}).
In summary, with implicit modeling,
we can learn the optimal latent code $\hat{\boldsymbol{z}}$ by
aggregating consistent (Eqs.~\eqref{c1}-\eqref{c2}) and complementary (Eq.~\eqref{c3}) information among multi-sweeps.

\subsection{Architecture Design of MV-DeepSDF} \label{arch}

To take advantage of multi-sweep consistency and complementarity in the latent feature space,
we design a new architecture for implicit modeling.
Our proposed architecture uses multi-sweep point clouds as model input and generates an optimal estimation of the 3D shape in the implicit space (a.k.a. the optimal latent code) as the output. The functionality of the network can be formulated as:
\begin{equation}
g_\alpha\left(\boldsymbol{Z}, \boldsymbol{P}\right)=\hat{\boldsymbol{z}}\rightarrow \boldsymbol{z}_{gt}\label{model},
\end{equation}
where $g(\cdot)$ denotes the desired network with associated parameters $\alpha$,
$\boldsymbol{Z}{=}\{\boldsymbol{z}_{1},\cdots, \boldsymbol{z}_{B} \}$ denotes the set of latent codes,
$B$ represents the number of observations or latent codes,
$\boldsymbol{P}{=}\{\mathcal{P}_{1},\cdots, \mathcal{P}_{B}\}$ denotes the set of multi-view point clouds,
and $\hat{\boldsymbol{z}} \in \mathbb{R}^{256}$ and $\boldsymbol{z}_{gt} \in \mathbb{R}^{256}$ represent
the estimated optimal and ground truth latent codes, respectively.
Hence, the goal is converted into learning a network to make $\hat{\boldsymbol{z}}$ approach $\boldsymbol{z}_{gt}$ for subsequent reconstructions.
To this end, we devise a new architecture to realize the functionality required by
simultaneously learning global features and latent codes for generating predicted latent codes.

\begin{figure}[t!]
	\centering
	\includegraphics[width=3.3in]{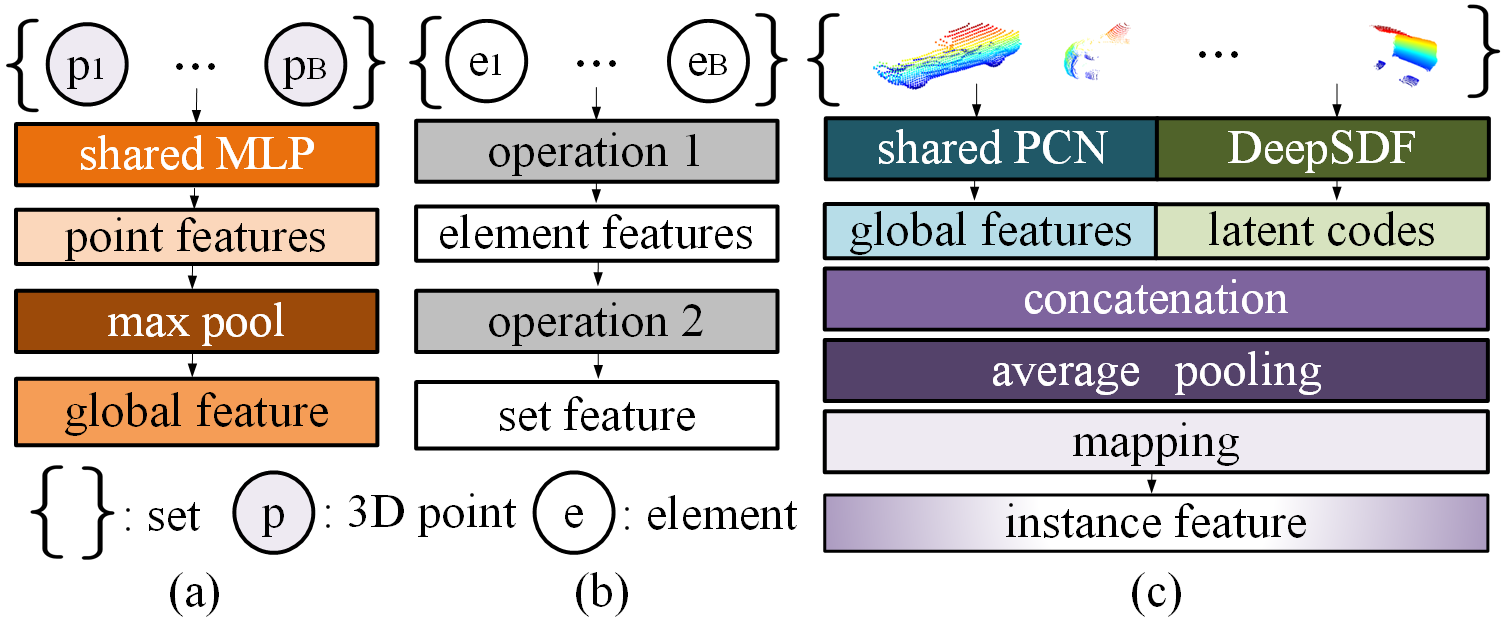}
	\caption{Comparison of three pipelines.
	(a) The pipeline of PointNet \cite{pointnet}.
	(b) The pipeline of an abstract generalized process to extract the set feature from the elements.
	(c) The proposed pipeline to extract the instance feature from multi-sweep point clouds.}
	\label{ele}
\end{figure}

Specifically, inspired by PointNet \cite{pointnet}, a pioneering work for 3D classification and segmentation,
we transform the problem of shape estimation in the implicit space into an element-to-set feature extraction problem.
Given a point cloud containing a set of 3D points,
where all 3D points consistently describe a single object since they all belong to the same instance, 
we can learn a global feature for each point cloud that aggregates complementary features from each individual 3D point.
However, only learning a global feature cannot be used for 3D reconstruction,
so we propose to abstract the pipeline of PointNet (see Figure \ref{ele}(a)) into a generalized process (see Figure \ref{ele}(b)).
As shown in Figure \ref{ele}(a), PointNet first computes 3D point features using a shared MLP on each 3D point
and then transforms these point features into a single global feature through max pooling.
Now, suppose there exists a set composed of multiple elements.
The goal of the generalized pipeline is to extract a feature for the entire set by aggregating all elements. To this end, we propose two generalized operations.
As shown in Figure \ref{ele}(b),
we define operation 1 as the computation of the element features (\eg, the shared MLP in PointNet)
and operation 2 as the transformation of the element features into a set feature (\eg, the max pooling operation in PointNet).

Using this generalized process, we can now devise our architecture to
simultaneously learn global features and latent codes for generating predicted latent codes.
%
%
First, we construct a global feature extractor (see yellow block in Figure \ref{network})
for learning a global feature for each point cloud that aggregates complementary features from each individual 3D point.
The architecture of this feature extractor is inspired by the PCN encoder \cite{pcn}, a variant of PointNet \cite{pointnet}.
This block is a stack of four PointNet \cite{pointnet} encoders with 128, 256, 512, and 1024 units, respectively.
Since this architecture is inspired by the PCN encoder \cite{pcn} and its global feature extraction operation
is applied to every individual point cloud, we name it \emph{shared PCN}.
Meanwhile, to exploit the implicit information of each element,
we employ a pre-trained DeepSDF \cite{deepsdf} model (as denoted by Eq.~\eqref{recon})
to generate a latent code for each partial point cloud.
Next, we concatenate the global features and latent codes to generate the element-level representations.
Then, we employ average pooling and mapping
to aggregate the element-level representations into a set-level predicted latent code.
As shown in Figure \ref{ele}(c), operation 1 of our pipeline consists of a global feature extractor and a pre-trained DeepSDF model, while operation 2 consists of concatenation, average pooling, and mapping.
This pipeline realizes a symmetric operation on unordered multi-sweep point clouds.

\subsection{Model Training of MV-DeepSDF} \label{train}
We adopt a curriculum learning strategy for model training, which consists of two stages.

\paragraph{Stage One.}
In stage one, we pre-train the DeepSDF decoder using watertight CAD models belonging to the car taxonomy of the ShapeNetV2 dataset \cite{shapenet}.
The latent code $\boldsymbol{z}$ and parameters of the decoder $\theta$ are jointly optimized as:
\begin{equation}
\underset{\theta,\left\{\boldsymbol{z}_j\right\}_{j=1}^J}{\arg \min } \sum_{j=1}^J\left(\frac{1}{\sigma^2}\left\|\boldsymbol{z}_j\right\|_2^2+\sum_{k=1}^K \mathcal{L}_{1}\left(f_\theta\left(\boldsymbol{z}_j, \boldsymbol{x}_k\right), s_k\right)\right),
\end{equation}
where $J$ represents the number of 3D shapes used for training and $K$ represents the number of points for each shape.
During this stage, the decoder gains prior knowledge of vehicle shapes and once trained, the decoder remains fixed for the latent code generation step of our pipeline.
Please refer to \cite{deepsdf} for the details of the training at this step.

\begin{figure}[t!]
	\centering
	\includegraphics[width=3.0in]{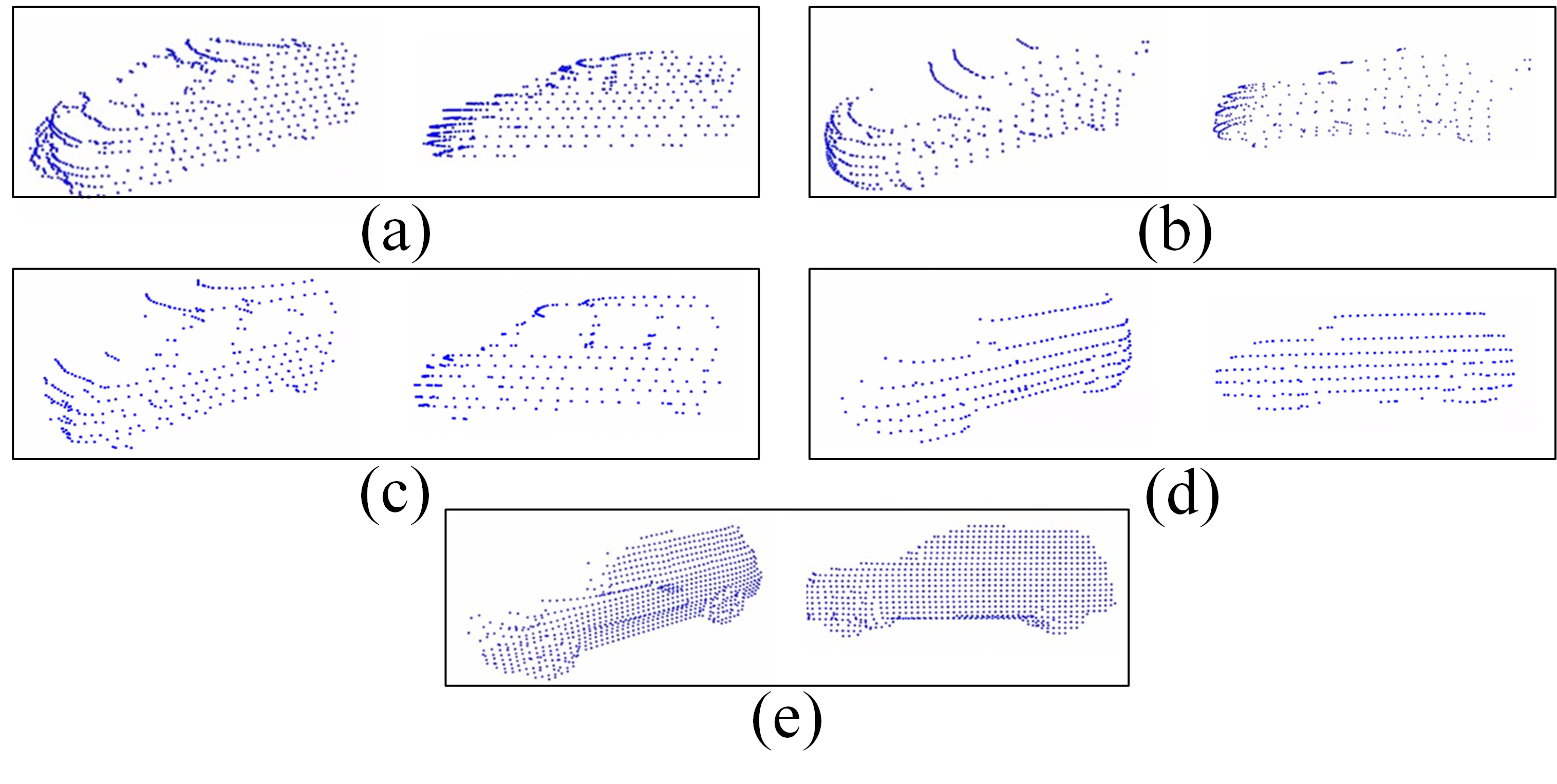}
	\caption{Visual comparison of in-the-wild partial point clouds and raycasted point clouds.
	(a) A partial point cloud from the Waymo tracking dataset \cite{waymo}.
	(b) The raycasted point cloud generated by PCGen \cite{pcgen} using the LiDAR parameters of Waymo.
	(c) A partial point cloud from the KITTI tracking dataset \cite{kitti}.
	(d) The raycasted point cloud generated by PCGen \cite{pcgen} using the LiDAR parameters of KITTI.
	(e) The raycasted point cloud obtained from the virtual depth camera approach used by DeepSDF \cite{deepsdf}.}
	\label{domain}
\end{figure}

\paragraph{Training Dataset Preparation for Stage Two.}
As shown in Eq.~\eqref{model}, ground truth latent codes, partial point clouds,
and their corresponding latent codes are required for training the reconstruction network.
However, real-world datasets, such as Waymo~\cite{waymo} and KITTI~\cite{kitti}, do not contain ground truth shapes,
so it is necessary to use a synthetic dataset \cite{shapenet} for generating the training data.
To resolve this problem, we utilize the latent codes of the training shapes
(\eg, the complete dense point cloud shown in Case 1 of Figure \ref{different}) as ground truth latent codes
in the second stage of model training.
As for the partial point clouds, the domain gap between training instances and in-the-wild instances directly determines
the ability of our network to generalize to real-world datasets.
To reduce the domain gap, we adopt PCGen \cite{pcgen} as our partial point cloud generation method.
PCGen places a virtual LiDAR with real-world parameters (resolution and sampling pattern) around the vehicle
and simulates the point cloud captured by the virtual LiDAR.
This differs from the method used by DeepSDF \cite{deepsdf}, where a simulated depth camera is used as the virtual sensor instead of a LiDAR.
A visual comparison of the real-world point cloud and simulated partial point clouds obtained using various methods is given in Figure \ref{domain}.
As shown in Figure \ref{domain}, the partial point clouds of PCGen \cite{pcgen} are visually closer to the in-the-wild LiDAR sweeps of Waymo \cite{waymo} and KITTI \cite{kitti} than those of the virtual depth camera approach \cite{deepsdf},
which only raycasts evenly in the inclination and azimuth directions.
Apart from this visual superiority, the advantages of generating partial point clouds with PCGen can also be observed
through the improved generalization ability of our network. 
See experiments for further details.

When generating training data from ShapeNetV2 \cite{shapenet}, we randomly sample one side of the vehicle to accurately reflect the
real-world scan of a LiDAR on an ego-vehicle.
In this way, we produce six partial point clouds for each vehicle.
The virtual LiDAR pose for each partial point cloud is generated using a set of restrictions.
These restrictions are set by considering the possible relative poses of an ego vehicle's LiDAR in the coordinate system of the other vehicle.
In particular, $\theta {\in} [0^{\circ},180^{\circ}]$ or $\theta {\in} [-180^{\circ},0^{\circ}]$, $ r{\in} [3,15]$, and $h {\in} [0.8,1.2]$
are used, where $\theta$ represents the azimuth, $r$ represents the distance between the LiDAR and the vehicle,
and $h$ represents the height of the LiDAR to the ground.
$\theta, r,$ and $h$ describes the pose of the virtual LiDAR with respect to the frame of the other vehicle.
Both $r$ and $h$ are expressed in the normalized space, where the size of the vehicle is normalized into the range $[-1,1]$.
Finally, we generate a latent code for each partial point cloud using Eq.~\eqref{recon}.

\paragraph{Stage Two.}
In stage two, the network is trained to reconstruct 3D vehicles from multi-view partial point clouds and their corresponding latent codes.
Consider the $c$th vehicle instance.
Let $\boldsymbol{P}_{B,c}$ represent the set of partial point clouds,
$\boldsymbol{Z}_{B,c}$ represent the set of latent codes corresponding to the multi-view point clouds,
and $\boldsymbol{z}_{gt,c}$ represent the ground truth latent code acquired from stage one. Furthermore, let
$g(\cdot)$ represent the function of the implicit shape prediction network (refer to yellow and red blocks in Figure \ref{network}) and $\alpha$ represent the parameters of the model.
The training objective of stage two can be defined as:
\begin{equation}
\underset{\alpha}{\arg \min } \sum_{c=1}^C \mathcal{L}_{2} \left( 
g_\alpha(\boldsymbol{Z}_{B,c},\boldsymbol{P}_{B,c}),\boldsymbol{z}_{gt,c}\right) \label{twin},
\end{equation}
where $C$ represents the number of instances employed for training and $\mathcal{L}_{2}$ is the Mean Squared Error loss.
The training details are presented in Sec. \ref{4.0}.

\begin{figure*}[t!]
	\centering
	\includegraphics[width=17.5cm]{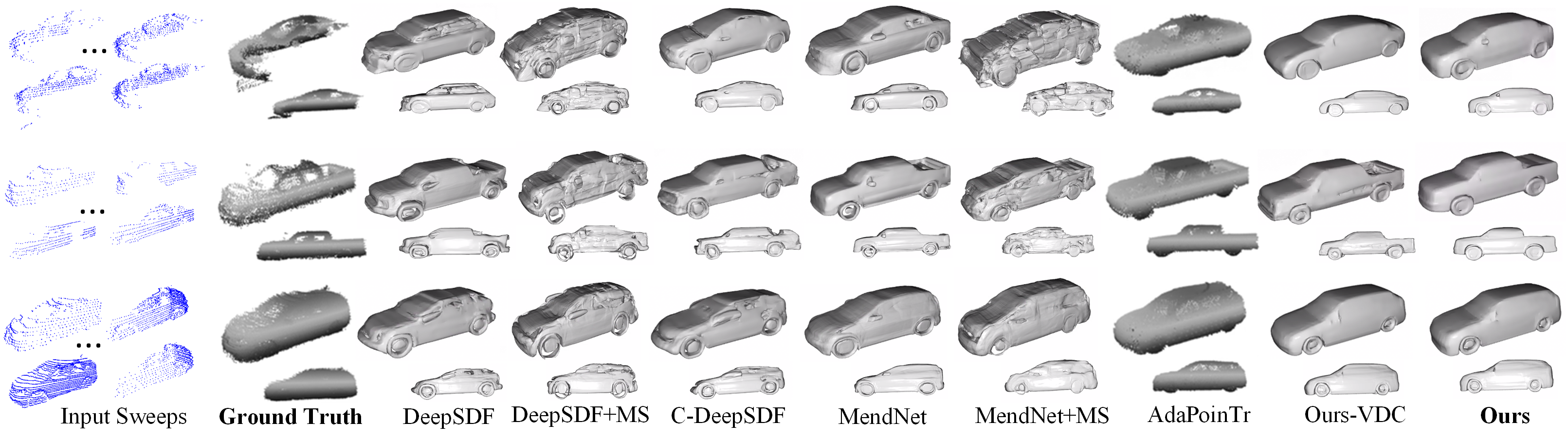}
	\caption{Visual comparison with the state-of-the-art methods (DeepSDF \cite{deepsdf}, C-DeepSDF \cite{csdf}, MendNet \cite{mending}, and AdaPoinTr \cite{adapointr}) on the Waymo \cite{waymo} dataset. DeepSDF+MS and MendNet+MS indicate the models with multi-sweep input.}
	\label{waymopic}
\end{figure*}

\section{Experiments} \label{exp}
In this section, we present qualitative and quantitative results on two real-world autonomous driving datasets,
namely Waymo \cite{waymo} and KITTI \cite{kitti}.
The multi-sweep point clouds for Waymo are collected from 136 unique vehicle instances of Waymo Open Dataset's tracking data \cite{waymo},
while those for KITTI are extracted from 233 vehicle instances of KITTI's tracking dataset \cite{kitti}.
Due to page limitations, experiment details and some results (\eg, the results on the synthetic dataset ShapeNetV2 \cite{shapenet})
are discussed in the supplementary material.

\subsection{Implementation Details} \label{4.0}
\paragraph{Architecture Details.} Our global feature extractor, named \textit{shared PCN}, is a variant of the PCN encoder \cite{pcn}. The embedded shared MLP layers are identical to those of PointNet \cite{pointnet}, which are implemented using 1D convolution layers. Moreover, since DeepSDF \cite{deepsdf} normalizes the latent codes into the range $[-1,1]$, we add a tanh layer after the last shared MLP to normalize the global features into the same range as the latent codes. The decoder is identical to that of DeepSDF \cite{deepsdf}. %
We report results based on an implementation with Python and PyTorch,
but our approach also supports implementation with MindSpore \cite{mindspore}.
\paragraph{Training Details.} In the first stage of training, we follow the same method as presented in DeepSDF \cite{deepsdf} to train the decoder using watertight CAD models from the car taxonomy of ShapeNetV2 \cite{shapenet}. Once trained, the decoder is fixed and projects a 256-dimensional latent code to an SDF in 3D space. In the second stage, we train the model for 20 epochs using the Adam optimizer \cite{adam} with a learning rate of 1e-5. The batch size is set to 1, allowing the model to simultaneously see all 6 frames of the given instance at each iteration. Since both the output of the model and the supervision are 1D vectors, the loss computation is very fast. It only takes 10 minutes to train our network on a single NVIDIA GeForce RTX 2080 GPU during stage two. However, the DeepSDF decoder \cite{deepsdf} used to prepare the training dataset for stage two through latent code generation is a more time-consuming process.

\subsection{Evaluation Protocol} \label{4.1}
\paragraph{Metrics.}
Since an in-the-wild instance does not have a ground truth 3D shape,
we follow \cite{mending} to evaluate the reconstruction results.
In particular, the multiple LiDAR sweeps are stacked to construct the ground truth points set $\boldsymbol{X}$.
Moreover, we sample 30,000 points on the surface of each reconstructed mesh to generate the reconstruction points set $\boldsymbol{Y}$.
We employ Asymmetric Chamfer Distance (ACD) \cite{mending},
which is the sum of the squared distance of each ground truth point to the nearest point in the reconstructed point set,
to evaluate reconstruction results on real-world datasets:
\begin{equation}
\operatorname{ACD}\left(\boldsymbol{X}, \boldsymbol{Y}\right)=\sum_{x \in \boldsymbol{X}} \min _{y \in \boldsymbol{Y}}\|x-y\|_2^2.\label{acd}
\end{equation} 
\par
For ACD, a smaller value is preferred.
In addition, we compute the recall of the ground truth points from the reconstructed shape, which is defined as:
\begin{equation}
\operatorname{Recall}(\boldsymbol{X}, \boldsymbol{Y})=\frac{1}{|\boldsymbol{X}|} \sum_{x \in \boldsymbol{X}}\left[\min _{y \in \boldsymbol{Y}}\|x-y\|_2^2<=t\right],
\end{equation}
where the threshold $t$ is set as 0.1, following \cite{mending}.

\paragraph{Competitors.}
We compare our approach with four state-of-the-art methods, including
three implicit modeling-based methods (\textbf{DeepSDF} \cite{deepsdf},
\textbf{C-DeepSDF} \cite{csdf}, and \textbf{MendNet} \cite{mending})
and \textbf{AdaPoinTr} \cite{adapointr}, which is built upon PoinTr++ \cite{pointr},
the winner of Multi-View Partial Point Cloud Challenge 2021 on Completion and Registration \cite{pointr++}.
Training data for DeepSDF and C-DeepSDF are generated following the original papers,
while MendNet uses PCGen \cite{pcgen} to generate training data.
The pipeline of AdaPoinTr \cite{adapointr} requires online generation of partial point clouds using the virtual depth camera approach and since it is not trivial to replace this process with PCGen, AdaPoinTr is trained using the default setup without PCGen.
Note that with multi-sweep point clouds, competitors generate multiple reconstruction results (one for each partial point cloud),
so we compare with the best single-shot reconstruction result,
which is the mesh with the minimum ACD among the multiple single-shot reconstructed meshes.
Furthermore, we report the results of DeepSDF and MendNet with multi-sweep inputs
(denoted as \textbf{DeepSDF+MS} and \textbf{MendNet+MS}),
in which we stack the partial point clouds of multi-sweeps into a single point cloud and perform single-shot-based reconstruction on this stacked point cloud.
For our approach,
in addition to the default setup which generates the training dataset using PCGen (denoted as \textbf{Ours}),
we also adapt our approach to use training data generated by
the virtual depth camera approach \cite{deepsdf} (denoted as \textbf{Ours-VDC}).

\begin{table}[t!]
	\centering
	\resizebox{0.99\columnwidth}{!}{
		\begin{tabular}{c|c|c|c}
			\hline\hline
				Method $\backslash$ Metric& $\mathrm{ACD}_{mean}$ $\downarrow$& $\mathrm{ACD}_{median}$ $\downarrow$& Recall $\uparrow$ \\ \hline
DeepSDF \cite{deepsdf}  & 6.26 & 5.81 & 93.51  \\ \hline
DeepSDF+MS  & 5.12 & 5.09 & 95.57  \\ \hline
C-DeepSDF \cite{csdf} & 6.21 & 5.64 & 93.98 \\ \hline
	MendNet \cite{mending} & 4.92 & 4.79 & 95.39 \\ \hline
  MendNet+MS & 4.85 & 4.77 & 95.76 \\ \hline
	AdaPoinTr$^\ast$\cite{adapointr} & 4.79 & 4.74 & 95.95 \\ \hline
Ours-VDC  & 4.76 & 4.55 & 96.05\\ \hline
\textbf{Ours}  & \textbf{3.36} & \textbf{2.26} & \textbf{96.84}\\ \hline
			
		\end{tabular}
	}
\caption{Comparison with the state-of-the-art methods on Waymo.
ACD $\downarrow$ is multiplied by $10^{3}$. Recall $\uparrow$ is presented as a percentage.
$^\ast$AdaPoinTr is trained with the default setup without PCGen on the large-scale ShapeNet-55~\cite{pointr} following the original paper~\cite{adapointr},
and then fine-tuned on the car taxonomy of ShapeNetV2 in our experiments. Other methods are only trained with the car taxonomy of ShapeNetV2. } 
		\label{tab2}
\end{table}

\begin{figure*}[t!]
	\centering
	\includegraphics[width=17.5cm]{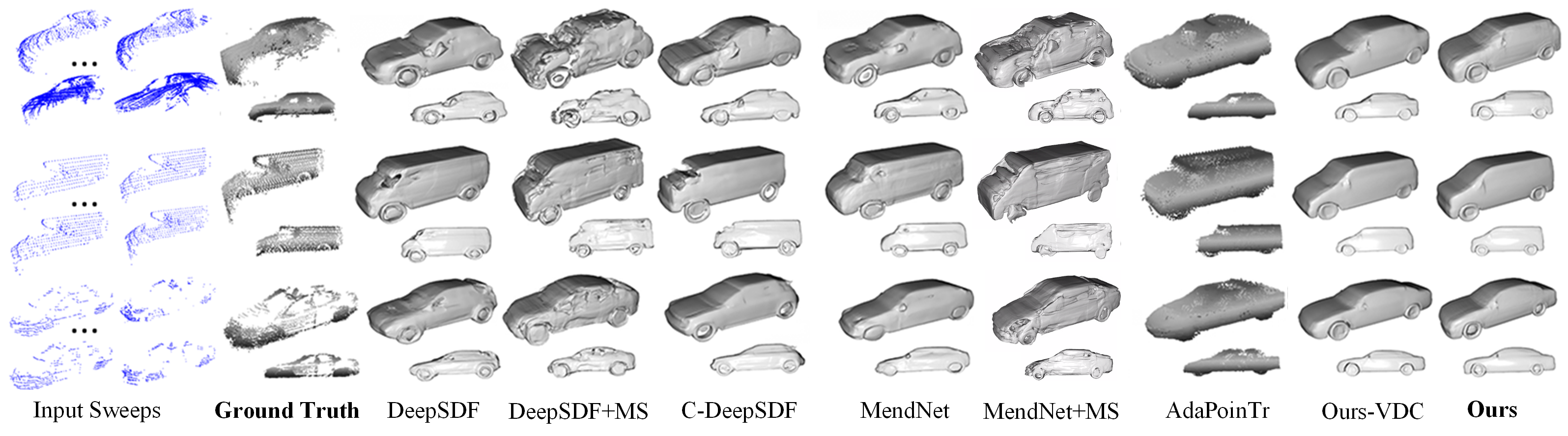}
	\caption{Visual comparison with the state-of-the-art methods (DeepSDF \cite{deepsdf}, C-DeepSDF \cite{csdf}, MendNet \cite{mending}, and AdaPoinTr \cite{adapointr}) on the KITTI \cite{kitti} dataset. DeepSDF+MS and MendNet+MS indicate the models with multi-sweep input.}
	\label{kittipic}
\end{figure*}

\subsection{Results on Waymo} \label{waymo}

The qualitative and quantitative comparisons of our approach with the state-of-the-art methods are presented in Figure \ref{waymopic} and Table \ref{tab2}, respectively.
Overall, our approach achieves the best reconstruction results both qualitatively and quantitatively.
Specifically, from Figure \ref{waymopic} and Table \ref{tab2},
we can see that DeepSDF \cite{deepsdf} and C-DeepSDF \cite{csdf} are sensitive to noise, resulting in low fidelity.
MendNet \cite{mending} generates more stable results than DeepSDF and C-DeepSDF, but its reconstruction meshes contain some irregular holes on the surface (see sixth column of Figure \ref{waymopic}).
Although AdaPoinTr \cite{adapointr} outputs point clouds with good fidelity, it expresses the shape with a limited resolution and thus fails to describe continuous local details (see seventh column of Figure \ref{waymopic}). \par
Compared with vanilla DeepSDF and MendNet, DeepSDF+MS and MendNet+MS yield messier surfaces/structures
(see fourth and eighth columns of Figure \ref{waymopic}), despite their better metric scores. 
This shows that even though transforming all partial point clouds into a single frame does benefit information aggregation, it also accumulates noise and annotation errors in the process. 
Hence, due to noise in real-world data, it is not practical to geometrically stack multi-sweep point clouds and
directly perform single-shot-based reconstruction. 
In comparison, our approach generates smooth watertight shapes (see last two columns of Figure \ref{waymopic})
with high fidelity (see Table \ref{tab2}).
Furthermore, Ours outperforms Ours-VDC, which indicates that using PCGen to prepare training partial point clouds
(refer to Figure \ref{domain}) reduces the domain gap between training and in-the-wild instances.

			

\begin{table}[t!]
	\centering
	\resizebox{0.99\columnwidth}{!}{
		\begin{tabular}{c|c|c|c}
			\hline\hline
				Method $\backslash$ Metric & $\mathrm{ACD}_{mean}$ $\downarrow$& $\mathrm{ACD}_{median}$ $\downarrow$& Recall $\uparrow$ \\ \hline
DeepSDF \cite{deepsdf}  & 6.81 & 6.17 & 80.65  \\ \hline
DeepSDF+MS   & 6.11 & 5.83 & 82.73  \\ \hline
C-DeepSDF \cite{csdf} & 6.73 & 5.99 & 80.77 \\ \hline
	MendNet \cite{mending} & 5.94 & 5.64 & 83.84 \\ \hline
  MendNet+MS & 5.83 & 5.61 & 84.26 \\ \hline
	AdaPoinTr \cite{adapointr} & 5.89 & 5.67 & 84.20 \\ \hline

Ours-VDC  & 5.75 & 5.24 & 84.39\\ \hline
\textbf{Ours}  & \textbf{4.27} & \textbf{3.01} & \textbf{85.88} \\ \hline
			
		\end{tabular}
	}
\caption{Comparison with the state-of-the-art methods on KITTI. ACD $\downarrow$ is multiplied by $10^{3}$. Recall $\uparrow$ is presented as a percentage. }
		\label{tab3}
\end{table}

\subsection{Results on KITTI}\label{kitti} 
The qualitative and quantitative comparisons of our approach against the state-of-the-art methods are presented in Figure \ref{kittipic} and Table \ref{tab3}, respectively.
From these results, we can also observe the superior performance of our approach over the state-of-the-art competitors.
Specifically, qualitatively, as shown in Figure \ref{kittipic}, our approaches (Ours and Ours-VDC)
are more robust to noise when compared to DeepSDF \cite{deepsdf}, DeepSDF+MS, and C-DeepSDF \cite{csdf}
and generate smoother, watertight surfaces compared to MendNet \cite{mending}, AdaPoinTr \cite{adapointr}, and MendNet+MS.
Quantitatively, as shown in Table \ref{tab3}, Ours yields the best ACD and Recall results,
while Ours-VDC performs the second best.

\subsection{Ablation Study} \label{ab}

To verify the effectiveness of the main components of our approach, we carry out a series of experiments on Waymo \cite{waymo} as shown in Table \ref{tab4}.
The first row of Table \ref{tab4} refers to the baseline model of our approach, which is a shared PCN encoder (Enc.) (yellow block in Figure \ref{network}) followed by an average pooling layer (Avg.).
This architecture is similar to that of MendNet \cite{mending}, which adds two stacked PointNet encoders \cite{pointnet} to the DeepSDF decoder.
In the second row, we directly use bitwise multiplication to merge the global features (B$\times$256, instead of B$\times$1024) and the latent codes (B$\times$256).
However, this results in obvious performance degradation,
which shows that directly merging the global features and latent codes obtained from multi-sweep information does not bring improvement to our baseline model.
In comparison, in the third row, we add our proposed components, namely the input of latent codes (Dep.), concatenation (Con.), and mapping (Map.), to the baseline model, but with max pooling.
This brings significant performance improvement compared to the first and second rows.
Then, in the last row, we further replace max pooling with average pooling, which includes all proposed components of this work.
This yields the best results.

\begin{table}[t!]
		\centering
			\resizebox{0.99\columnwidth}{!}{
			\begin{tabular}{ccccc|cc}
				\hline\hline
				\multicolumn{2}{c}{Operation 1}          & \multicolumn{3}{c|}{Operation 2} &\multicolumn{2}{c}{Metric} \\ \cline{1-7}
	Enc. & Dep.  & Mer. & Pool  & Map.  &   ACD $\downarrow$ & Recall $\uparrow$                \\ \hline  
	\ding{51} & &   & Avg. &&4.89& 95.50 \\ 
	\ding{51} &\ding{51}  &  Mul. & Avg. &&7.33& 84.93\\
	\ding{51} &\ding{51} & Con. & Max.& \ding{51} &4.35& 96.61 \\
	\ding{51} &\ding{51} & Con.  & Avg.& \ding{51} &3.36& 96.84 \\ \hline
				
			\end{tabular}
			}
	\caption{The ablation study of the proposed framework. Refer to Figure \ref{ele}(b) to see the definitions of operations 1 and 2.
	Enc.: the shared PCN encoder;
	Dep.: latent code generation through DeepSDF as input;
	Mer.: the operation to merge global features and latent codes;
	Map.: the mapping conducted by the fully-connected layer;
	Con.: the concatenation;
	Mul.: the bitwise multiplication;
	Avg.: average pooling;
	Max.: max pooling.
	ACD $\downarrow$ is multiplied by $10^{3}$. Recall $\uparrow$ is presented as a percentage.}
			\label{tab4}
	\end{table}
\subsection{Extension to Other Taxonomies}
\begin{figure}[h!]
	\centering
	\includegraphics[width=3.0in]{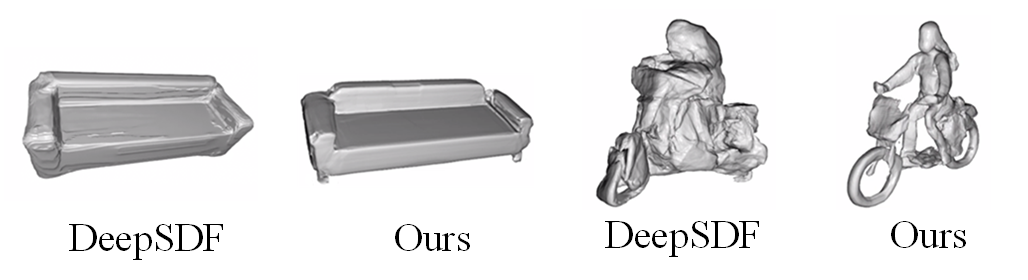}
	\caption{Visual comparison of DeepSDF and MV-DeepSDF on the indoor sofa of ShapeNetV2 and the outdoor cyclist of Waymo.
	}\label{cyc}
\end{figure}

It is straightforward to extend the proposed shape completion framework to other taxonomies. Suppose that we want to extend MV-DeepSDF to perform shape completion on sofas. First, a new DeepSDF decoder \cite{deepsdf} needs to be pre-trained using watertight CAD models from the sofa taxonomy of ShapeNetV2 \cite{shapenet} or another synthetic dataset. Then, PCGen \cite{pcgen} or another sampling technique (\eg., the approaches proposed in \cite{deepsdf,pointr}) can be used to generate partial point clouds for each sofa instance. These partial point clouds are then passed into the pre-trained DeepSDF decoder \cite{deepsdf} to produce their corresponding latent codes. Finally, the partial point clouds and latent codes can be used together to train MV-DeepSDF. Following the pipeline above, we evaluate on indoor sofas of ShapeNetV2 and outdoor cyclists of Waymo and show that our method is versatile and extendible to other taxonomies (see Figure \ref{cyc}).

While existing datasets, such as ScanNet \cite{scannet} and Semantic3D \cite{semantic3d}, provide real-world point cloud data for many taxonomies, the availability of tracking data for multi-sweep scans is still fairly limited to the autonomous driving industry. Hence, it is necessary to first obtain the corresponding multi-sweep tracking labels for the desired class when choosing to adopt MV-DeepSDF on a new taxonomy.
 
\section{Conclusion}\label{con}
In this work, we propose a new MV-DeepSDF framework to facilitate implicit modeling with multi-sweep point clouds for autonomous driving. The main idea is to abstract 3D vehicle reconstruction from multi-sweeps into an element-to-set feature extraction problem. Namely, we consider the multi-sweeps of a vehicle as elements composing a set and infer the set feature, which is an optimal estimation of the 3D shape described in the abstract implicit space. In particular, we present a theoretical analysis of multi-sweep consistency and complementarity in the latent feature space. Guided by this analysis, we design a new architecture to optimally estimate the Signed Distance Function shape of a vehicle from its in-the-wild multi-sweep point clouds.
Qualitative and quantitative evaluations on both real-world and synthetic datasets
show the superiority of our approach over the state-of-the-art methods.
\par
\noindent\textbf{Limitation.}
Despite promising results, our approach still relies on a synthetic 3D dataset to gain prior knowledge of 3D shapes for reconstruction.
Exploring reconstruction by learning directly from real data is worthy of further study.
\section*{Acknowledgments}
 The authors gratefully acknowledge the support of MindSpore, CANN, and Ascend AI Processor. The authors would also like to thank Andrew Yang and Hunter Schofield for their assistance.


\clearpage

\section* {Supplementary Material}

\subsection*{A. Overview}
This material provides quantitative and qualitative experimental results, dataset and implementation details, and discussions that are supplementary to the main paper. 

\subsection*{B. Dataset Details} \label{sh}
The multi-sweep LiDAR point clouds for Waymo are collected from 136 unique vehicle instances of Waymo Open Dataset’s tracking data \cite{waymo}, while those for KITTI are extracted from 233 unique vehicle instances of KITTI’s tracking dataset \cite{kitti}. 
In total, we extracted 3943 partial point clouds from the 136 vehicle instances of Waymo and 4235 partial point clouds from the 233 vehicle instances of KITTI.
These raw multi-sweep point clouds are directly used as model input to obtain experimental results on our model and the state-of-the-art methods \cite{deepsdf, csdf, mending, adapointr}. When performing inference, all partial point cloud frames from the given instance are simultaneously passed into our model as input.
\par To construct the ground truth stacked point cloud, we first aggregate all partial point clouds within a multi-sweep to generate a dense stacked point cloud. Since this stacked point cloud contains unwanted noise, such as ground plane points and points lying on the exterior or interior of the vehicle surface, we perform statistical outlier removal on the stacked point cloud, which computes the average distance of a point from its neighbours and removes all points lying farther away from their neighbours than average. Denoising is essential for dataset processing since the presence of noise in the ground truth shape can result in false positives and false negatives in model performance, whereby a messy shape generated by a model that fits to the noise of the ground truth is deemed high fidelity and a smooth shape generated by a noise-robust model is deemed low fidelity.
\subsection*{C. Visualization Results}
\begin{figure*}[t!]
	\centering
	\includegraphics[width=15.5cm]{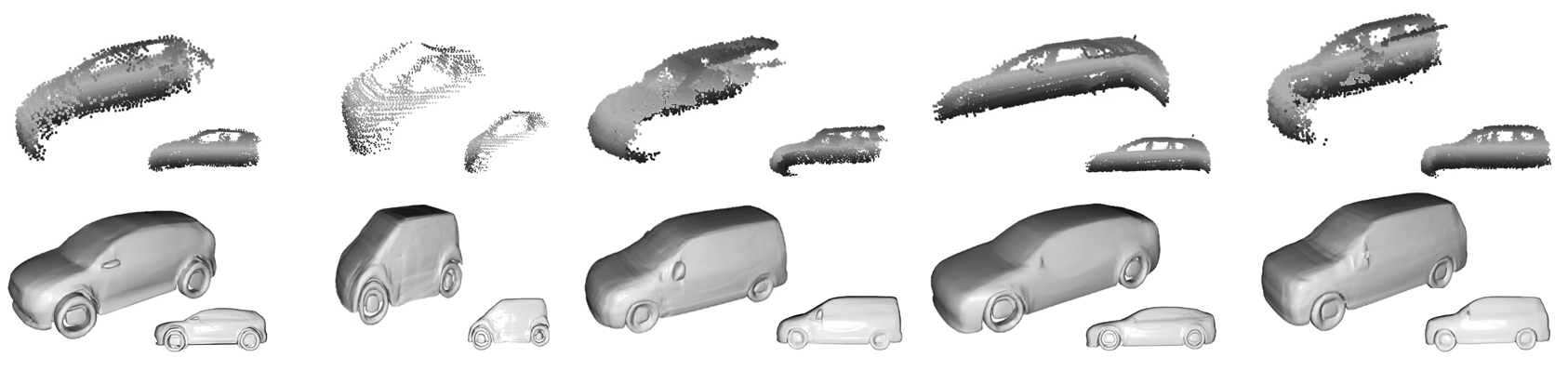}
	\caption{Additional visualization results of MV-DeepSDF on the Waymo \cite{waymo} and KITTI \cite{kitti} datasets.}
	\label{more}
\end{figure*}

\begin{figure*}[t!]
	\centering
	\includegraphics[width=15.5cm]{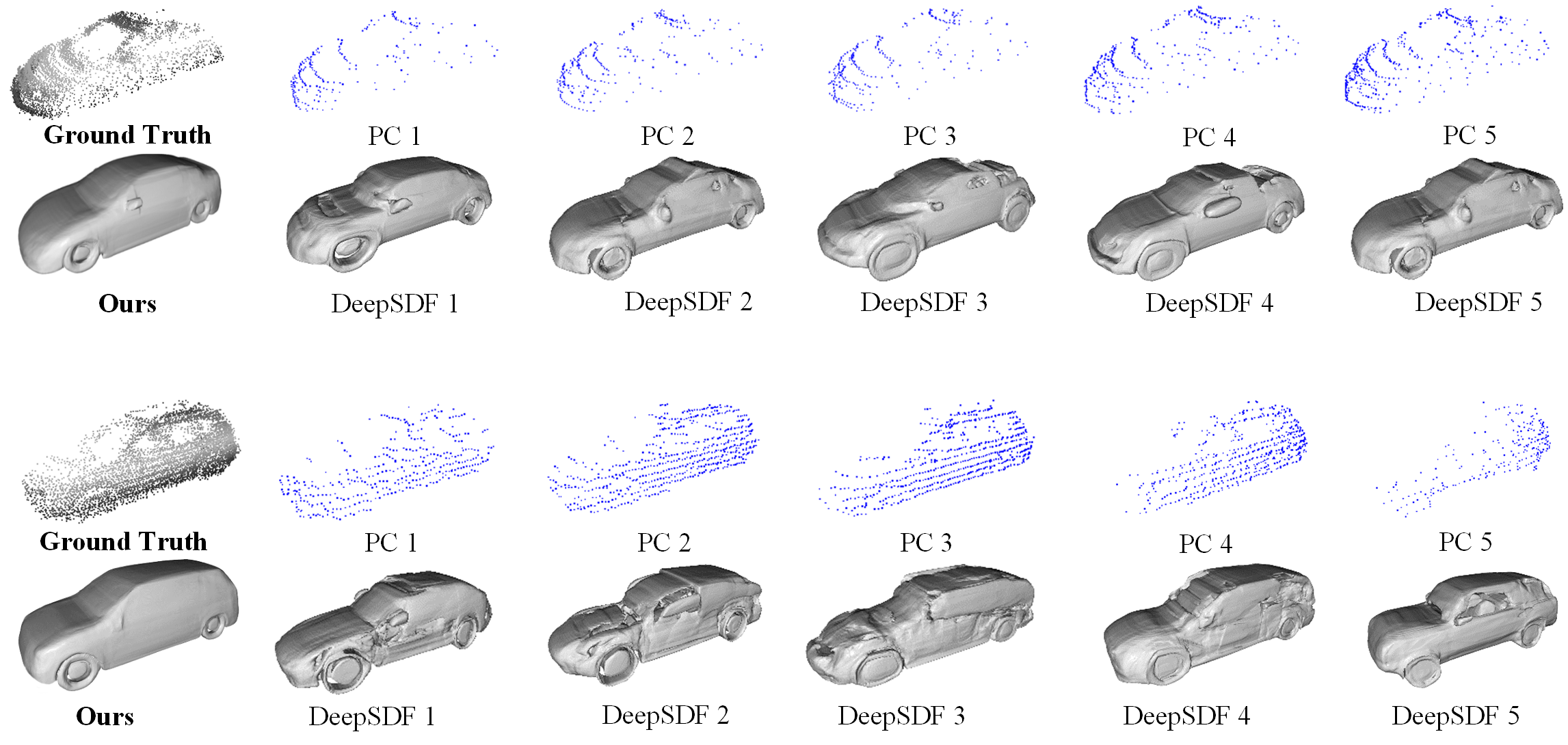}
	\caption{Visual comparison of MV-DeepSDF and DeepSDF \cite{deepsdf} on Waymo \cite{waymo}. Individual point clouds are given in the first row and their corresponding reconstruction results from vanilla DeepSDF in the second row.}
	\label{more2}
\end{figure*}

Due to the page limitation of the main paper, we present more visualization results of our model in Figure \ref{more}. Additionally, to evidently present the significant improvement of our method over the baseline vanilla DeepSDF \cite{deepsdf}, a visual comparison on Waymo \cite{waymo} is given in Figure \ref{more2}.
\subsection*{D. Results on ShapeNetV2} 
We randomly preserve 300 vehicles from the car taxonomy of ShapeNetV2 \cite{shapenet} as the test dataset. The remainder is used to train our network in stage one. Each vehicle instance is comprised of 6 partial point clouds generated by PCGen \cite{pcgen} under Waymo's LiDAR parameters. Note that the following results are generated solely using PCGen \cite {pcgen} to sample partial point clouds and the use of other sampling techniques (\eg., the approaches proposed in \cite{deepsdf,pointr}) would yield different results on the same ShapeNetV2 dataset \cite{shapenet}.
\begin{figure*}[h!]
	\centering
	\includegraphics[width=17.6cm]{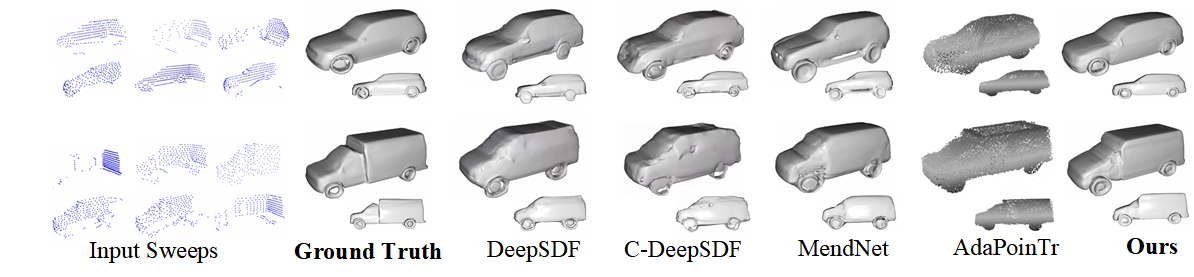}
	\caption{Visual comparison with the state-of-the-art methods (DeepSDF \cite{deepsdf}, C-DeepSDF \cite{csdf}, MendNet \cite{mending}, and AdaPoinTr \cite{adapointr}) on the ShapeNetV2 \cite{shapenet} dataset.}
	\label{shapenet}
\end{figure*}
\begin{figure}[thpb]
	\centering
	\includegraphics[width=3.3in]{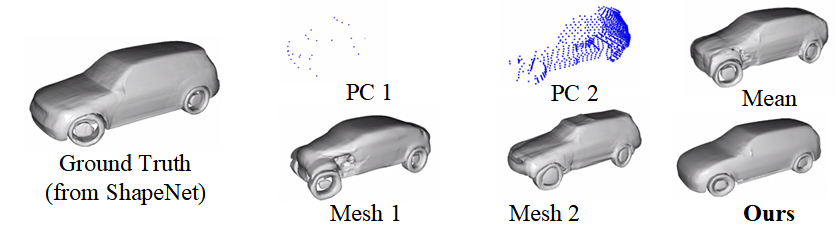}
	\caption{Comparison of the result of our approach to that of the mean latent code on ShapeNetV2 \cite{shapenet}. The proposed network is not fine-tuned. }
	\label{non}
\end{figure}

\paragraph{Metrics.}  Since ground truth shapes are readily available in synthetic datasets such as ShapeNetV2 \cite{shapenet}, we use Chamfer Distance (CD) \cite{cd} to evaluate the 3D reconstruction results. Following DeepSDF \cite{deepsdf}, we sample 30,000 points on the surface of both the ground truth and reconstructed mesh. Given two point sets, the CD is the sum of the squared distance of each point to the nearest point in the other point set: 

\begin{equation}
CD\left(\boldsymbol{X}, \boldsymbol{Y}\right)=\sum_{x \in \boldsymbol{X}} \min _{y \in \boldsymbol{Y}}\|x-y\|_2^2+\sum_{y \in \boldsymbol{Y}} \min _{x \in \boldsymbol{X}}\|x-y\|_2^2. \label{cd}
\end{equation} \par
As outlined in the main paper, we only compare the results of our model with the best single-shot reconstruction result,
which is the mesh with the minimum CD among the multiple single-shot reconstructed meshes.

\paragraph{Qualitative and Quantitative Comparison.}
 The qualitative and quantitative comparison of our approach against the state-of-the-art methods (DeepSDF \cite{deepsdf}, C-DeepSDF \cite{csdf}, MendNet \cite{mending}, and AdaPoinTr \cite{adapointr}) are presented in Figure \ref{shapenet} and Table \ref{tab1}, respectively. Note that when testing on ShapeNetV2 \cite{shapenet}, since PCGen \cite{pcgen} is used to generate both the training and test dataset, we only present Ours, in contrast with the comparison of Ours with Ours-VDC in the main paper. The meshes generated by DeepSDF \cite{deepsdf}, C-DeepSDF \cite{csdf}, and MendNet \cite{mending} show high fidelity compared to their performance on real-world datasets, but still show inferior performance to Ours. AdaPoinTr \cite{adapointr} also produces shapes with decent fidelity, but the reconstructed result is not watertight and expresses the shape with a limited resolution which fails to describe the continuous surface of the vehicle.

\begin{table}[htbp]
	\centering
		\begin{tabular}{c|c|c}
			\hline\hline
			Method $\backslash$ Metric & $\mathrm{CD}_{mean}$ $\downarrow$& $\mathrm{CD}_{median}$ $\downarrow$\\ \hline
DeepSDF \cite{deepsdf} & 5.47 & 5.15\\ \hline
C-DeepSDF \cite{csdf} & 5.31 & 5.03   \\ \hline
MendNet \cite{mending} & 4.22 & 3.65  \\ \hline
AdaPoinTr \cite{adapointr} & 4.10 & 3.36  \\ \hline
\textbf{Ours} & \textbf{3.17} & \textbf{2.54}  \\ \hline
			
		\end{tabular}
\caption{Comparison of the proposed network with the state-of-the-art approaches on ShapeNetV2 \cite{shapenet}. CD is multiplied by $10^{3}$. }
		\label{tab1}
\end{table}\par

\subsection*{E. Comparison to a Non-Learning Approach}
We now present an alternative non-learning approach, computing the mean latent code, for the task of multi-sweep 3D vehicle reconstruction. As introduced in \cite{deepsdf}, linear interpolation between two latent codes in the latent space can also generate meaningful shape representations. Moreover, averaging is a common method of linear interpolation used for reducing error among multi-observation data. To this end, we investigate the effect of computing the mean latent code from the single-shot-based latent codes of a given multi-sweep and using this mean latent code for mesh reconstruction.  We present the case shown in Figure \ref{non}, where two single-shot partial point clouds, PC 1 and PC 2, are used to generate two latent codes, $\boldsymbol{z}_{1}$ and $\boldsymbol{z}_{2}$, and meshes, Mesh 1 and Mesh 2, respectively. We define the mean latent code as $\boldsymbol{z}_{mean}=0.5(\boldsymbol{z}_{1}+\boldsymbol{z}_{2})$ and generate the corresponding mesh, denoted by Mean. As shown, Mean is simply a uniform fusion of Mesh 1 and Mesh 2. Moreover, Mean is inferior to Mesh 2, the best single-shot in this example, which is also inferior to Ours, the result of our proposed model.


 \begin{table}[htbp]
	\centering
		\begin{tabular}{c|c|c}
			\hline\hline
			 Num of PCs & $\mathrm{ACD}_{mean}$$\downarrow$ & $\mathrm{ACD}_{median}$$\downarrow$  \\ \hline
3  & 3.47 & 2.44   \\ \hline
6  & 3.36 & 2.26  \\ \hline
9  & 3.32 & 2.21  \\ \hline

		\end{tabular}
\caption{Ablation study on Waymo \cite{waymo} using different numbers of point clouds per instance. ACD is multiplied by $10^{3}$.  }
		\label{tab5}
\end{table}

\subsection*{F. Effect of Number of Point Clouds}
The number of frames corresponding to an individual vehicle instance in Waymo \cite{waymo} and KITTI \cite{kitti} ranges up to 240 partial point clouds per instance. However, the vast majority of instances only contain between 3 to 9 partial point clouds. In this section, we investigate the relationship between the number of partial point clouds provided for each instance during stage two of training and overall model performance. Table \ref{tab5} presents the experimental results of providing different numbers of partial point clouds to our model on Waymo \cite{waymo}. As shown, our model performance improves as the number of point clouds increases. However, since generating the latent code for each partial point cloud with DeepSDF \cite{deepsdf} is a timely process (around 10 seconds), we choose 6 observations per instance as a trade-off between performance and efficiency. 

\subsection*{G. Effect of Number of Points Per Point Cloud}
 The number of points captured in a single frame of Waymo \cite{waymo} and KITTI \cite{kitti} mostly falls into a range of 300 to 1000 points. Thus, we set the number of points per point cloud as 256 in our framework for performing FPS. In this section, we investigate the relationship between the number of points per point cloud during inference and model performance. Table \ref{tab6} presents the experimental results of varying the number of points per point cloud on both DeepSDF and our model with Waymo \cite{waymo}. As shown, when the number of points decreases, the performance of DeepSDF drops dramatically whereas our method holds steady.

\begin{table}[h!]

	\centering
	\resizebox{0.99\columnwidth}{!}{
		\begin{tabular}{c|c|c|c|c}
			\hline\hline
    Num of Points & \multicolumn{2}{|c|}{256}  & \multicolumn{2}{|c}{128}  \\ \hline
				 Metric& $\mathrm{ACD}_{mean}$ $\downarrow$ & $\mathrm{ACD}_{median}$ $\downarrow$ &
				 $\mathrm{ACD}_{mean}$ $\downarrow$ & $\mathrm{ACD}_{median}$ $\downarrow$ \\ \hline

	DeepSDF & 6.26 & 5.81 & 12.52 & 8.51 \\ \hline
	\textbf{Ours} & \textbf{3.36} & \textbf{2.26} & \textbf{3.47}  & \textbf{2.64}\\ \hline
			
		\end{tabular}
	}
\caption{Ablation study on Waymo \cite{waymo} using different numbers of points per point cloud. ACD is multiplied by $10^{3}$. }
		\label{tab6}
\end{table}



{\small
\bibliographystyle{ieee_fullname}
\bibliography{egbib}
}

\end{document}